\documentclass{article}
\usepackage{ijcai25}

\usepackage{times}
\usepackage{soul}
\usepackage{url}
\usepackage[hidelinks]{hyperref}
\usepackage[utf8]{inputenc}
\usepackage[small]{caption}
\usepackage{graphicx}
\usepackage{amsmath}
\usepackage{amsthm}
\usepackage{booktabs}
\usepackage{algorithm}
\usepackage{algorithmic}
\usepackage[switch]{lineno}

\usepackage{natbib}
\usepackage{cite}
\usepackage{microtype}
\usepackage[usenames,svgnames]{xcolor}
\usepackage{setspace}
\usepackage{booktabs}
\usepackage{mydef}
\usepackage{booktabs}
\usepackage{tabularx}
\usepackage{framed}
\usepackage{bbding}
\usepackage{subfloat}
\usepackage{adjustbox}
\usepackage{multirow}
\usepackage{pifont}
\usepackage{multirow}

\title{Connector-S: A Survey of Connectors in Multi-modal Large Language Models}

\author{
Xun Zhu~\textsuperscript{1, \dag} \and
Zheng Zhang~\textsuperscript{1,\dag} \and
Xi Chen~\textsuperscript{1, \dag} \and
Yiming Shi~\textsuperscript{1} \and
Miao Li~\textsuperscript{1, \ding{41}}\And
Ji Wu~\textsuperscript{1, 2, 3, 4, \ding{41}}
\affiliations
\textsuperscript{1} Department of Electronic Engineering, Tsinghua University \quad
\textsuperscript{2} College of AI, Tsinghua University \\
\textsuperscript{3} Beijing National Research Center for Information Science and Technology \\
\textsuperscript{4} Center for Big Data and Clinical Research, Institute for Precision Medicine, Tsinghua University \\
\dag\ Equal Contribution \qquad
\ding{41} Corresponding Author
\emails
\{zhu-x24, zzhang24, chenxi24, sym23\}@mails.tsinghua.edu.cn,
\{miao-li, wuji\_ee\}@tsinghua.edu.cn
}

\begin{document}

\maketitle

\begin{abstract} 
With the rapid advancements in multi-modal large language models (MLLMs), connectors play a pivotal role in bridging diverse modalities and enhancing model performance.
However, the design and evolution of connectors have not been comprehensively analyzed, leaving gaps in understanding how these components function and hindering the development of more powerful connectors.
In this survey, we systematically review the current progress of connectors in MLLMs and present a structured taxonomy that categorizes connectors into atomic operations (mapping, compression, mixture of experts) and holistic designs (multi-layer, multi-encoder, multi-modal scenarios), highlighting their technical contributions and advancements. 
Furthermore, we discuss several promising research frontiers and challenges, including high-resolution input, dynamic compression, guide information selection, combination strategy, and interpretability.
This survey is intended to serve as a foundational reference and a clear roadmap for researchers, providing valuable insights into the design and optimization of next-generation connectors to enhance the performance and adaptability of MLLMs.

\end{abstract}

\section{Introduction}
With the remarkable advancements in large language models (LLMs) propelling progress towards general-purpose AI, there has been a significant growing focus on extending these models to multi-modal domains, leading to the development of multi-model large language models (MLLMs).
The current mainstream MLLM architectures follow a similar paradigm~\citep{zhang2024mm}:
the modality encoder, which compresses raw information, such as image or audio, into a more compact representation;
the connector, which alleviates the gap between modalities, thus facilitating the adaptation of multi-modal inputs to LLMs;
the LLM backbone, which generates the text responses in an auto-regressive manner;
and the generator, which is optional for generating more modalities output besides text.

Rather than training from scratch, a common and logical approach is to start with a pre-trained encoder and a pre-trained LLM, aiming to enhance the efficacy of modality representation and mitigate computational expenses of LLM pre-training, respectively~\citep{yin2023survey}.
In the above paradigm, the connector plays a crucial role in aligning the multi-modal inputs into a consistent form and space~\citep{cha2024honeybee}:
On the one hand, the connector bridges modalities and the language model by translating modality features into tokens that the language model can understand.
The quality of the conveyed tokens directly impacts the overall performance of the MLLM.
On the other hand, compared to the fixed architectures of encoders and LLMs, the connector is more lightweight and can be easily improved, offering greater flexibility and efficiency.

In one of the pioneering works in MLLMs, LLaVA~\citep{liu2024visual}, the connector is merely a simple linear layer that provides basic mapping functionality.
As the demand for more powerful capabilities and the complexity of scenarios increased, advanced connectors incorporate various mechanisms, such as improved mapping~\citep{liu2024improved}, compression based on spatial relation~\citep{chen2023minigpt} or semantic perception~\citep{li2023blip}, and mixture of experts (MoE)~\citep{li2025cumo} to better handle the intricacies of multi-modal inputs.
In addition, Dense Connector~\citep{yao2024dense} and Uni-Med~\citep{zhu2024uni} have optimized multi-layer feature utilization and multi-task joint learning conflicts by exploring connector design, respectively.
The evolution reflects increasing recognition of the key role of connectors in improving the overall performance of MLLM and adaptability to various complex scenarios.

\paragraph{Motivations.}
With the widespread application of MLLMs in various domains, the design of the connector is constantly updated to meet the increasing demands for more powerful capabilities and complex scenarios.
\citet{song2023bridge} has explored modality alignment methods for MLLMs, due to time constraints, many methods and scenarios have not yet appeared and been discussed.
Several studies~\citep{yin2023survey,zhang2024mm} have conducted surveys on the entire flow of MLLMs, while offering relatively brief and insufficient coverage of connectors.
The lack of a comprehensive survey on connectors makes it hard for researchers to establish a clear cognition of the underlying mechanisms of different connectors, thereby impeding the development of next-generation connectors.
Therefore, we propose a more fine-grained taxonomy to systematically review and summarize the current status of connectors in MLLMs.

\definecolor{natureblue}{RGB}{0, 114, 189}
\definecolor{naturegreen}{RGB}{119, 172, 48}
\definecolor{natureorange}{RGB}{217, 83, 25}
\definecolor{naturepurple}{RGB}{126, 47, 142}
\definecolor{natureyellow}{RGB}{237, 177, 32}

\tikzstyle{leaf}=[draw=none,
    rounded corners, minimum height=1em,
    fill=naturegreen!40, text opacity=1,
    fill opacity=.5, text=black, font=\scriptsize,
    inner xsep=3pt,
    inner ysep=1pt,
    text centered, 
]

\tikzstyle{middle}=[draw=none,
    rounded corners, minimum height=1em,
    fill=natureblue!40, text opacity=1,
    fill opacity=.5, text=black, font=\scriptsize,
    inner xsep=3pt,
    inner ysep=1pt,
    text centered, 
]

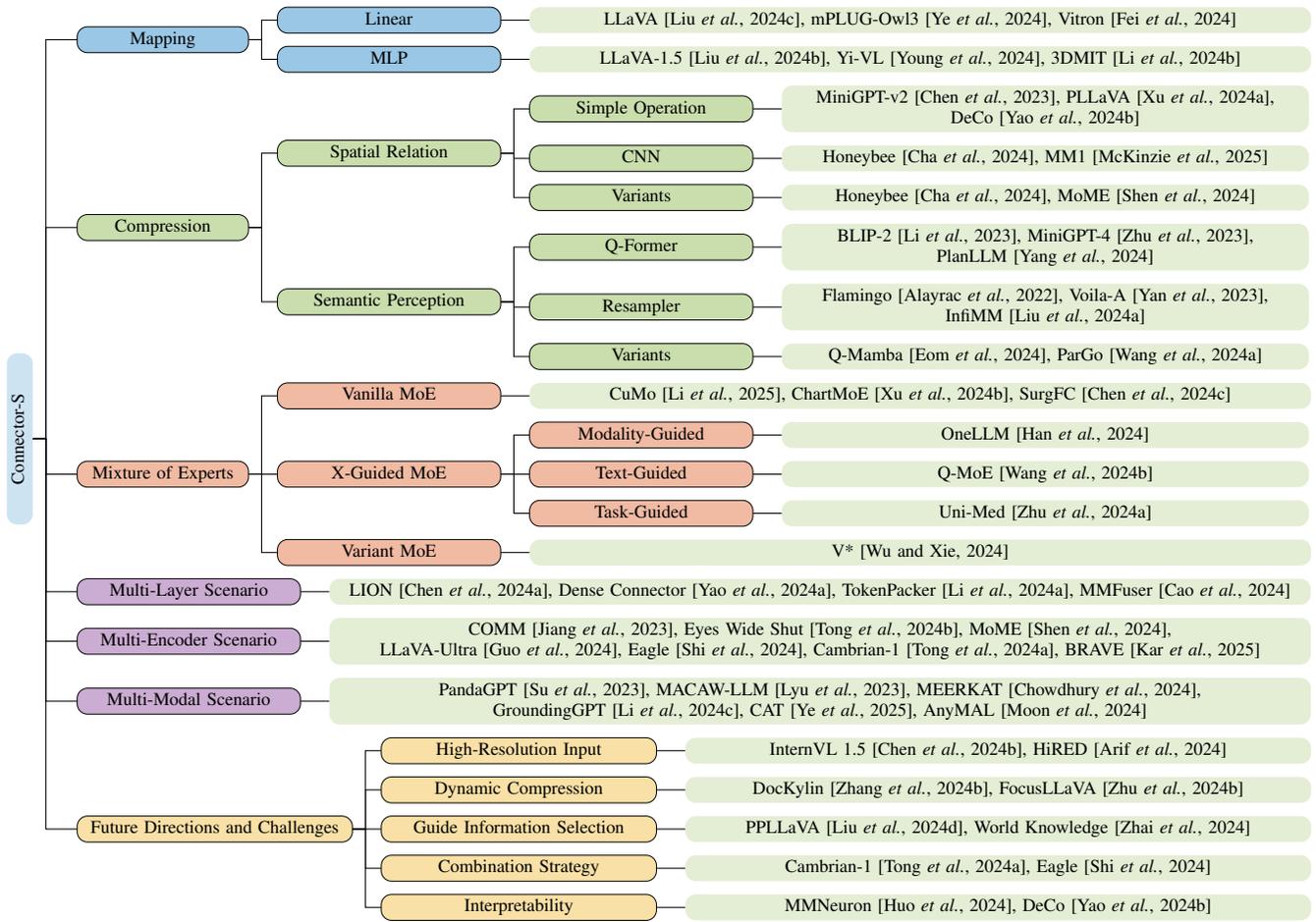
\begin{figure*}[ht]
\centering
\begin{forest}
  for tree={
    forked edges,
    grow=east,
    reversed=true,
    anchor=base west,
    parent anchor=east,
    child anchor=west,
    base=middle,
    font=\scriptsize,
    rectangle,
    line width=0.1pt,
    draw=black,
    rounded corners,
    align=center,
    text centered, 
    minimum width=2em,
    s sep=5pt,
    inner xsep=3pt,
    inner ysep=1pt,
  },
  where level=1{text width=4.5em, text centered}{}, 
  where level=2{text width=6em, text centered}{},
  where level=3{text centered}{},
  where level=4{text centered}{},
  where level=5{text centered}{},
  [Connector-S, middle, rotate=90, anchor=north, edge=black, text width=6em
    [Mapping, fill=natureblue!40, edge=black, text width=6em
        [Linear, fill=natureblue!40, text width=8em, edge=black
            [LLaVA \citep{liu2024visual}{,} mPLUG-Owl3 \citep{ye2024mplug}{,} Vitron \citep{fei2024vitron}, leaf, text width=29.5em, edge=black]
        ]
        [MLP, fill=natureblue!40, text width=8em, edge=black
            [LLaVA-1.5 \citep{liu2024improved}{,} Yi-VL \citep{young2024yi}{,}  3DMIT~\citep{li20243dmit}, leaf, text width=29.5em, edge=black]
        ]
    ]
    [Compression, fill=naturegreen!40, edge=black, text width=6em
        [Spatial Relation, fill=naturegreen!40, text width=8em, edge=black
            [Simple Operation, fill=naturegreen!40, text width=8em, edge=black
                [MiniGPT-v2 \citep{chen2023minigpt}{,} PLLaVA \citep{xu2024pllava}{,}\\ DeCo \citep{yao2024deco}, leaf, text width=19.7em, edge=black]
            ]
            [CNN, fill=naturegreen!40, text width=8em, edge=black
                [Honeybee \citep{cha2024honeybee}{,} MM1 \citep{mckinzie2025mm1}, leaf, text width=19.7em, edge=black]
            ]
            [Variants, fill=naturegreen!40, text width=8em, edge=black
                [Honeybee \citep{cha2024honeybee}{,} MoME \citep{shen2024mome}, leaf, text width=19.7em, edge=black]
            ]
        ]
        [Semantic Perception, fill=naturegreen!40, text width=8em, edge=black
            [Q-Former, fill=naturegreen!40, text width=8em, edge=black
                [BLIP-2 \citep{li2023blip}{,} MiniGPT-4 \citep{zhu2023minigpt}{,}\\ PlanLLM \citep{yang2024planllm}, leaf, text width=19.7em, edge=black]
            ]
            [Resampler, fill=naturegreen!40, text width=8em, edge=black
                [Flamingo \citep{alayrac2022flamingo}{,} Voila-A \citep{yan2023voila}{,}\\ InfiMM \citep{liu2024infimm}, leaf, text width=19.7em, edge=black]
            ]
            [Variants, fill=naturegreen!40, text width=8em, edge=black
                [Q-Mamba \citep{eom2024query}{,} ParGo \citep{wang2024pargo}, leaf, text width=19.7em, edge=black]
            ]
        ]
    ]
    [Mixture of Experts, fill=natureorange!40, edge=black, text width=6em
        [Vanilla MoE, fill=natureorange!40, text width=8em, edge=black
            [CuMo \citep{li2025cumo}{,} ChartMoE \citep{xu2024chartmoe}{,} SurgFC \citep{chen2024surgfc}, leaf, text width=29.5em, edge=black]
        ]
        [X-Guided MoE, fill=natureorange!40, text width=8em, edge=black
            [Modality-Guided, fill=natureorange!40, text width=8em, edge=black
                [OneLLM \citep{han2024onellm}, leaf, text width=19.7em, edge=black]
            ]
            [Text-Guided, fill=natureorange!40, text width=8em, edge=black
                [Q-MoE \citep{wang2024q}, leaf, text width=19.7em, edge=black]
            ]
            [Task-Guided, fill=natureorange!40, text width=8em, edge=black
                [Uni-Med \citep{zhu2024uni}, leaf, text width=19.7em, edge=black]
            ]
        ]
        [Variant MoE, fill=natureorange!40, text width=8em, edge=black
            [V* \citep{wu2024v}, leaf, text width=29.5em, edge=black]
        ]
    ]
    [Multi-Layer Scenario, fill=naturepurple!40, edge=black, text width=8em
            [LION \citep{chen2024lion}{,} Dense Connector \citep{yao2024dense}{,} TokenPacker \citep{li2024tokenpacker}{,} MMFuser \citep{cao2024mmfuser}, leaf, text width=37.2em, edge=black]
        ]
    [Multi-Encoder Scenario, fill=naturepurple!40, edge=black, text width=8em
            [COMM~\citep{jiang2023clip}{,} Eyes Wide Shut \citep{tong2024eyes}{,} MoME \citep{shen2024mome}{,}\\
            LLaVA-Ultra \citep{guo2024llava}{,} Eagle \citep{shi2024eagle}{,} Cambrian-1 \citep{tong2024cambrian}{,} BRAVE \citep{kar2025brave}, leaf, text width=37.2em, edge=black]
    ]
    [Multi-Modal Scenario, fill=naturepurple!40, edge=black, text width=8em
            [PandaGPT~\citep{su2023pandagpt}{,} MACAW-LLM \citep{lyu2023macaw}{,} 
            MEERKAT~\citep{chowdhury2024meerkat}{,} \\
            GroundingGPT \citep{li2024groundinggpt}{,}
            CAT~\citep{ye2025cat}{,} AnyMAL~\citep{moon2024anymal},
            leaf, text width=37.2em, edge=black]
        ]
    [
    Future Directions and Challenges, fill=natureyellow!40, edge=black, text width=10em
    [
    High-Resolution Input, fill=natureyellow!40, text width=10em, edge=black
    [
    InternVL 1.5 \citep{chen2024far}{,} HiRED \citep{arif2024hired}, leaf, text width=23.5em, edge=black
    ]
    ]
    [
    Dynamic Compression, fill=natureyellow!40, text width=10em, edge=black
    [
    DocKylin \citep{zhang2024dockylin}{,} FocusLLaVA \citep{zhu2024focusllava}, leaf, text width=23.5em, edge=black
    ]
    ]
    [
    Guide Information Selection, fill=natureyellow!40, text width=10em, edge=black
    [
    PPLLaVA \citep{liu2024ppllava}{,} World Knowledge \citep{zhai2024world}, leaf, text width=23.5em, edge=black
    ]
    ]
    [
    Combination Strategy, fill=natureyellow!40, text width=10em, edge=black
    [
    Cambrian-1 \citep{tong2024cambrian}{,} Eagle \citep{shi2024eagle}, leaf, text width=23.5em, edge=black
    ]
    ]
    [
    Interpretability, fill=natureyellow!40, text width=10em, edge=black
    [
    MMNeuron \citep{huo2024mmneuron}{,} DeCo \citep{yao2024deco}, leaf, text width=23.5em, edge=black
    ]
    ]
    ]
]
\end{forest}
\caption{A taxonomy of connectors in multi-modal large language models with representative examples.}
\label{f1}
\end{figure*}

\paragraph{Contributions.}
The contributions of this work can be summarized from the following three aspects:
(1) \textit{A structured taxonomy}.
A broad overview of the field is presented with a structured taxonomy that covers atomic operations and holistic designs.
(2) \textit{A comprehensive review}. 
Based on the proposed taxonomy, the current research progress of connectors in MLLMs is systematically delineated.
(3) \textit{Some research frontiers and directions}. 
We discuss promising research frontiers and point out possible future directions.

\section{Taxonomy}
To enhance our understanding of the dynamic evolution of connectors in MLLMs, we identify pivotal research endeavors, analyze their motivations, and succinctly encapsulate their primary technical contributions. 
As illustrated in Figure~\ref{f1}, this survey establishes a new taxonomy, which firstly examines the connector design of these works from two different perspectives, i.e., atomic operations in basic scenarios and holistic designs in complex scenarios.
Then, we briefly introduce these two perspectives as follows:
\begin{itemize}
\item \textbf{Atomic connector operations} refer to the basic components of MLLM connectors, which are designed as simple yet versatile units tailored to different functional requirements of basic scenarios.
By utilizing these atomic operations, connectors can achieve mapping, compression, and expert integration.
Furthermore, they can be combined to create more complex connectors, bridging the modality gap in a targeted and flexible way.
\item \textbf{Holistic connector designs} focus on the challenges of enhancing the capabilities of MLLMs in sophisticated scenarios from the multi-layer features, multi-encoder outputs to multi-modal contexts.
Effective holistic designs help MLLMs consider the interplay between different layers of visual features, combine insights from different visual encoders, and handle diverse modalities in a coherent and efficient manner.
\end{itemize}
In the following sections, we present a comprehensive survey along the two perspectives and corresponding categories of our taxonomy for connectors in MLLMs.
Due to some MLLMs possibly employing multiple methods in connectors, there may be instances of overlap among these models.
Finally, based on the previous summary, we open the discussion for challenges and opportunities of future connector research, including high-resolution input, dynamic compression, guide information selection, combination strategy, and interpretability.

\section{Atomic Connector Operations}
\subsection{Mapping}
Since the LLM backbones are primarily trained on generic text, there is an inevitable semantic gap exists when dealing with multi-modal features. 
Mapping operations first flatten 2D or 3D features into 1D in a specific order and directly align the dimension of representations from other modalities with textual token embeddings.

\paragraph{Linear.}
As the basic connector method, the simplicity of linear mapping makes it an attractive choice for initial feature transformation.
LLaVA~\citep{liu2024visual} uses a linear projection matrix to connect the representation produced by the visual encoder to the LLM backbone.
mPLUG-Owl3~\citep{ye2024mplug} still adopts a linear layer to further process long visual sequence inputs, while Vitron~\citep{fei2024vitron} utilizes three independent linear mappings to process image, video, and sketch features, respectively.

\paragraph{MLP.}
The multi-layer perceptron (MLP) consist of multiple linear layers connected by activation functions, introducing non-linearity into the feature transformation process.
LLaVA-1.5~\citep{liu2024improved} finds that changing from the original linear projection to MLP can improve LLaVA's multi-modal capabilities.
Yi-VL~\citep{young2024yi} and 3DMIT~\citep{li20243dmit} use a two-layer MLP and a three-layer MLP with layer normalizations to align 2D image features and 3D object features, respectively.

The primary advantages of mapping operations lie in the simplicity and lightweight nature, which facilitates fast convergence during alignment training.
However, mapping operations are limited in their ability to compress redundant information, which may lead to long token sequences that impose computational burden on the subsequent LLM backbone.

\subsection{Compression}
The complete set of tokens derived from other modalities encompasses both useful and redundant information.
To achieve the optimal balance between information representation and the number of tokens, it is crucial to implement efficient compression operations through the connector.
The current mainstream compression operations can be categorized into two main types: spatial relation and semantic perception.

\subsubsection{Spatial Relation}
The core idea behind spatial relation compression is rooted in observation: features from adjacent regions tend to be more similar in the original modality representation. 
Leveraging the spatial proximity, this category of compression operations can be further divided into several subcategories:

\paragraph{Simple Operation.}
Pooling and dimensionality reduction via token concatenation and projection both serve as simple operations to compress the number of tokens.
For example, MiniGPT-v2~\citep{chen2023minigpt} simply concatenates 4 adjacent visual tokens in the embedding space and projects them together into one single token with the textual embedding dimension.
PLLaVA~\citep{xu2024pllava} processes video features with the average pooling strategy to smooth the feature distribution along the temporal dimension.
DeCo~\citep{yao2024deco} compresses $N$ visual tokens by reshaping them into a 2D tensor of size $\left(N^{1/2}, N^{1/2}\right)$, applying adaptive average pooling to reduce the size to $\left(M^{1/2}, M^{1/2}\right)$, flattening into $M$ tokens, and projecting through a linear layer to match the textual embedding dimension.

\paragraph{CNN.}
Convolutional neural networks (CNNs) can preserve local spatial structures and enhance spatial understanding through hierarchical feature extraction.
For instance, Honeybee~\citep{cha2024honeybee} proposes C-Abstractor that consists of stacked ResNet blocks and adaptive pooling, injecting locality-aware design into the compression process.
MM1~\citep{mckinzie2025mm1} adopts C-Abstractor as the connector, verifying the effectiveness of the connector design compared with average pooling and attention pooling.

\paragraph{Variants.}
Considering that CNN introduces overly strict inductive biases for locality, Honeybee~\citep{cha2024honeybee} also presents D-Abstractor,
which enhances the locality-awareness while maintaining flexibility based on deformable attention. 
To avoid feature mismatch, MoME~\citep{shen2024mome} proposes adaptive deformable transformation, which combines adaptive average pooling and deformable attention to obtain compressed and self-enhanced visual features.

\subsubsection{Semantic Perception}
Features with similar semantics always tend to exhibit stronger correlations across modalities.
By focusing on semantic content rather than just spatial proximity, semantic perception compression aims to extract and preserve the most semantically relevant information while reducing redundancy.
Leveraging semantic understanding, this category can be further divided into several subcategories:

\paragraph{Q-Former.}
First proposed in BLIP-2 \citep{li2023blip}, Q-Former introduces a fixed number of learnable tokens as queries, which retrieve relevant information from other modality representations through cross-attention.
Since Q-Former involves more parameters and typically requires additional training, some studies use the pre-trained Q-Former to continue training for specific tasks. 
MiniGPT-4~\citep{zhu2023minigpt} utilizes the frozen Q-Former with a linear layer to enhance its visual understanding capabilities.
PlanLLM~\citep{yang2024planllm} promotes video procedure planning by continuing to train Q-Former, integrating sample-specific visual state information and textual step knowledge.

\paragraph{Resampler.}
First introduced in Flamingo~\citep{alayrac2022flamingo}, resampler
insert additional cross-attention layers between pre-trained and frozen LLM layers, whre the keys and values are obtained from the compressed vision features while the queries are derived from the language inputs.
Voila-A~\citep{yan2023voila} further adapts resampler by using the gaze heatmap features as the keys within the attention mechanism, optimizing feature expression.
InfiMM~\citep{liu2024infimm} keeps the resampler trainable in three-stage training strategies for vision-language alignment, visual question answering knowledge injection and unreshing conversation ability.

\paragraph{Variants.}
To reduce the computational and memory complexity of the transformer, Q-Mamba \citep{eom2024query} introduces a bidirectional Mamba layer to process visual features from a pre-trained vision encoder. These features are then projected into a fixed number of learnable queries using cross-attention, enhanced by a causal Mamba prior.
ParGo~\citep{wang2024pargo} employs learnable local and global tokens to extract information separately, enhancing regional relationships while controlling token count.

Compression operations help distill essential information while eliminating redundancy, thus improving both the efficiency and effectiveness of the model.
Spatial relation compression provides simple and direct, albeit somewhat crude methods for reducing token count.
Semantic perception compression offers deeper semantic understanding at the cost of increased complexity and training requirements.

\subsection{Mixture of Experts}
Mixture of Experts (MoE) is a powerful architectural paradigm composed of two key components: a set of experts, each specializing in distinct aspects of the multi-modal features, and a router that dynamically selects and combines their outputs according to the input guidance information.
With the shared goal of achieving optimal alignment of multi-modal features, compression operations focus on extracting a minimal number of tokens, while MoE aims to leverage the specialized representations of multiple experts.
MoE's ability to capture a wide range of features and patterns makes it particularly suitable for the complex and varied nature of multi-modal data. 

\subsubsection{Vanilla MoE}
Vanilla MoE refers to a standard MoE framework where each multi-modal token is independently processed by the router to compute expert assignment weights. 
CuMo~\citep{li2025cumo} employs four expert models, each consisting of a two-layer MLP. 
The router is a linear layer that computes expert weights based on each input token, applying a top-2 gating mechanism.
This means that only the top two experts with the highest weights are selected for each token, allowing the connector to focus on the most relevant information while maintaining computational efficiency.
ChartMoE~\citep{xu2024chartmoe} adopts four experts of the linear layer with a top-2 gating router, achieving distinct preferences and effective processes in the fused information of background, table, json and code.
To enhance surgical multi-modal understanding, SurgFC~\citep{chen2024surgfc} extends the vanilla MoE connector to the medical domain, utilizing MLP for both experts and router.

\subsubsection{X-Guided MoE}
X-guided MoE expands the input to the router beyond the individual multi-modal token itself. 
The improved router strategy of X-guided MoE can leverage additional information to make more informed routing decisions, thereby improving the connector's ability to dynamically allocate resources and select the most relevant experts. 

\paragraph{Modality-Guided.}
Modality information enables the connector to quickly classify tokens by their source modality, facilitating efficient expert allocation, especially in the case of input of multiple modalities.
For example, OneLLM~\citep{han2024onellm} addresses the challenge of aligning eight modalities, including image, audio, video, point cloud, depth map, normal map, IMU and fMRI brain activity, to language within a unified framework.
To facilitate modality switching, OneLLM introduces a set of learnable tokens for each modality.
When tokens from different modalities are inputted, their corresponding modality tokens are concatenated with the input tokens and fed into the router. 

\paragraph{Text-Guided.}
For connectors, understanding the textual information such as instructions or descriptions corresponding to other modality tokens helps them understand how to convert tokens according to requirements.
Q-MoE~\citep{wang2024q} introduces a text-guided MoE connector, which routes different query tokens to pre-defined task experts by making use of cross-attention between the text representation and output of each expert.

\paragraph{Task-Guided.}
In a unified multi-task learning framework, task-specific information provides critical context for the connector to classify tokens and determine the most suitable experts.
Uni-Med~\citep{zhu2024uni} introduces a unified medical foundation model that employs C-MoE, a well-designed router which calculates routing weights based on concatenated visual and task-specific tokens and activates different experts for each task, efficiently addressing the multi-task interference problem in MLLMs.

\subsubsection{Variant MoE}
Given the expectation that different experts will exhibit distinct preferences for different types of tokens, experts in the MoE framework can also be designed differently, which has promoted the development of variant MoE.
For example, the connector of V*~\citep{wu2024v} consists of two projection experts: a linear layer and a resampler.
And V* designs a search algorithm as the router to flexibly switch between these two projection modules:
1) none searched target: the linear layer;
2) one or two searched targets: the linear layer for searched targets, the resampler for global image;
3) more than two searched targets: the resampler.

The inherent variability of multi-modal data necessitate a flexible and adaptive approach to feature transmission and integration.
MoE provides this flexibility by dynamically allocating resources and selecting the most appropriate expert for different parts of the input, thereby enhancing the model's ability to handle diverse tasks and scenarios.

\section{Holistic Connector Designs}
Atomic connector operations have utilized single-layer features, typically the last or penultimate layer derived from the modality encoder, to explore and provide basic alignment functions.
In contrast, holistic connector designs are proposed for more complex scenarios which are essentially multi-dimensional and heterogeneous in features.
These scenarios arise from three main sources:
1) multi-layer features, extracted from various layers of the original encoder, each sharing the same dimensionality but offering different levels of abstraction;
2) multi-encoder features, obtained from different encoders, each with its own dimensionality and semantic focus;
3) multi-modal features, from additional modalities, further enriching the representation space.
The core challenge lies in effectively integrating these diverse features to align with the textual embedding space.

\begin{figure*}[!ht] 
\centerline{\includegraphics[scale=0.53]{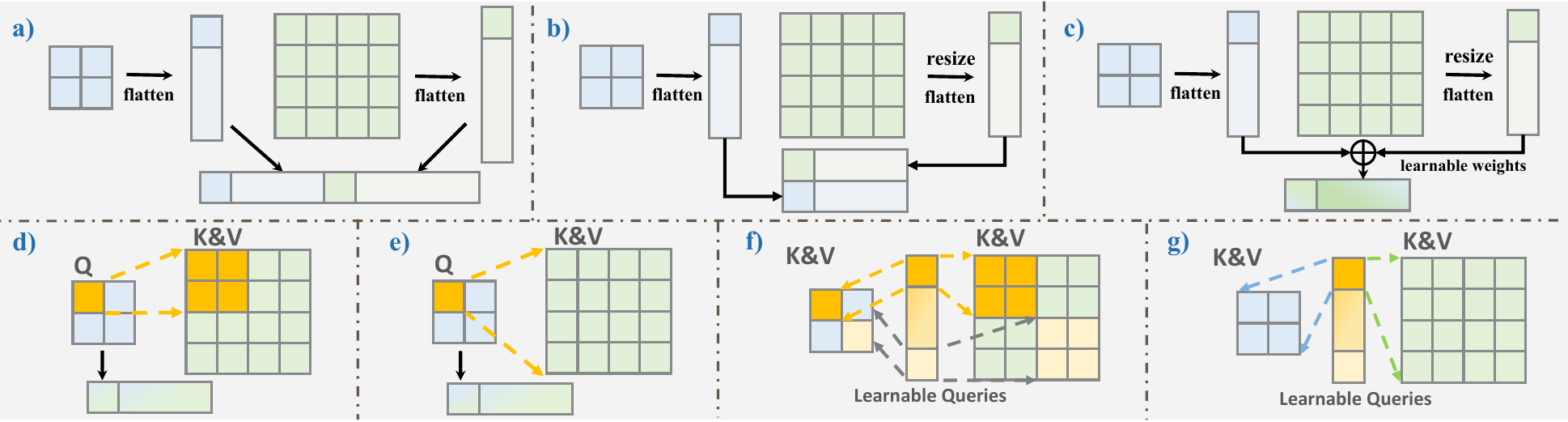}}
\caption{The schematic diagram of the fusion strategies: (a) token concatenation. (b) channel concatenation. (c) channel weighted addition. (d) local cross-attention. (e) global cross-attention. (f) local cross-attention (learnable queries). (g) global cross-attention (learnable queries).}
\label{f2}
\end{figure*}

\begin{table*}
\tiny
\tabcolsep= 0.5cm
\centering
\begin{tabular}{cccc}
\toprule
\textbf{Model} & \textbf{Visual Encoder} & \textbf{Multi-Layer Selection} & \textbf{Fusion Strategy} \\ 
\midrule
LION~\citep{chen2024lion} & EVA-CLIP-ViT-G & {[}16, 32, 47{]} & Global Cross-Attention→Token Concatenation\\
Dense Connector~\citep{yao2024dense} STI & CLIP-ViT-L & {[}8, 16, 24{]} & Token Concatenation \\
Dense Connector~\citep{yao2024dense} SCI & CLIP-ViT-L & {[}8, 16, 24{]} & Channel Concatenation \\
Dense Connector~\citep{yao2024dense} DCI & CLIP-ViT-L / SigLIP-ViT-SO & {[}1-12, 13-24{]} / {[}1-13, 14-26{]} & Weighted Channel Addition→Channel Concatenation \\
TokenPacker~\citep{li2024tokenpacker} & CLIP-ViT-L & {[}24{]} (Q) {[}12, 16, 22, 23{]} (K, V) & Local Cross-Attention \\
MMFuser~\citep{cao2024mmfuser} & CLIP-ViT-L & {[}23{]} (Q) {[}3, 8, 13, 18{]} (K, V) & Token Concatenation→Global Cross-Attention \\ \bottomrule
\end{tabular}
\caption{A summary of representative connectors in the multi-layer scenario according to the multi-layer selection and fusion strategy.}
\label{t1}
\end{table*}

\begin{table*}[!]
\tiny
\tabcolsep= 0.4cm
\centering
\begin{tabular}{ccc}
\toprule
\textbf{Model} & \textbf{Multi-Encoder Selection} & \textbf{Fusion Strategy} \\
\midrule
COMM~\citep{jiang2023clip} & CLIP, DINOv2 & Weighted Channel Addition→Token Concatenation \\
Eyes Wide Shut~\citep{tong2024eyes} A-MoF & CLIP-ViT-L, DINOv2-ViT-L & Weighted Channel Addition→Token Concatenation \\
Eyes Wide Shut~\citep{tong2024eyes} I-MoF &  CLIP-ViT-L, DINOv2-ViT-L & Token Concatenation \\
MoME~\citep{shen2024mome} & EVA-CLIP-G, DINOv2-ViT-L, Pix2Struct-Base & Weighted Channel Addition \\
LLaVA-Ultra~\citep{guo2024llava} & CLIP-ViT-L, SAM-ViT-L & Weighted Channel Addition\\ 
\multirow{2}{*}{Eagle~\citep{shi2024eagle}} & CLIP-ViT-L, SigLIP-ViT-SO400M, & \multirow{2}{*}{Channel Concatenation} \\
 & EVA-02-L-Det, SAM-ViT-L, DINOv2-ViT-L &  \\
\multirow{2}{*}{Cambrian-1~\citep{tong2024cambrian}} & CLIP, ConvNeXt-XXL, Pix2Struct-02-Large, & \multirow{2}{*}{Local Cross-Attention (Learnable Queries)} \\
 & ConvNeXt-XXL, DINOv2-ViT-L &  \\
BRAVE~\citep{kar2025brave} & EVA-CLIP-G, CLIP-ViT-L, SILC-ViT-G, ViT-e, DINOv2-ViT-L & Token Concatenation→Global Cross-Sttention (Learnable Queries) \\
\bottomrule
\end{tabular}
\caption{A summary of representative connectors in the multi-encoder scenario according to the multi-encoder selection and fusion strategy.}
\label{t2}
\end{table*}

We observe that existing popular fusion strategies, despite their variations in designs, can be broadly represented by the following several categories:
1) token concatenation, directly concatenating tokens from different layers or encoders to form a longer sequence;
2) channel concatenation, concatenating the tokens along the channel dimension, which may require resizing to ensure the number of tokens remains consistent;
3) channel weighted addition, using learnable weights to weight and sum features along the channel dimension without increasing both the sequence length and channel dimension;
4) local cross-attention, using features of specific layers as queries, dividing features of other layers into sub-regions as keys and values, and aggregating each query with the keys and values derived from its corresponding sub-region;
5) global cross-attention, using global features of specific layers as queries and the global features of other layers as keys and values for aggregation;
6) local cross-attention (learnable queries), where features of all extracted layers are divided into sub-regions respectively, and each query aggregates with the keys and values derived from its corresponding sub-regions.
Note that the number of sub-regions per layer must be divisible by the number of queries.
7) global cross-attention (learnable queries), where global features of all extracted layers are used as keys and values for aggregation.
As shown in Figure~\ref{f2}, we visualize the schematic diagram of the fusion strategies for easier understanding.

\begin{figure*}[!ht] 
\centerline{\includegraphics[scale=0.47]{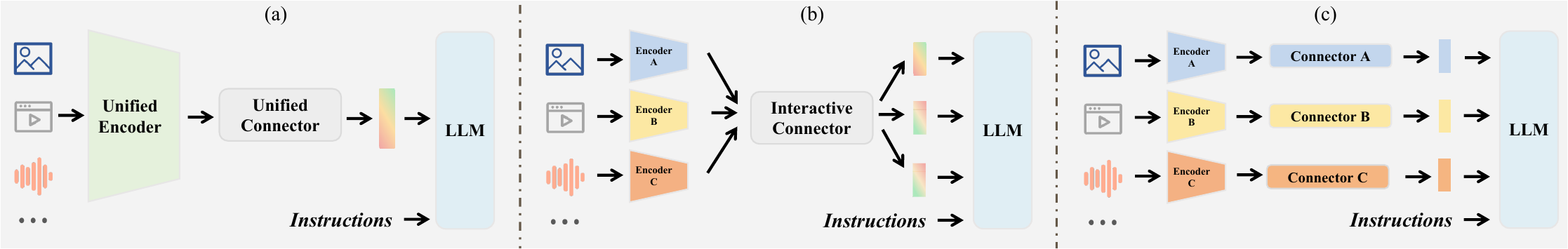}}
\caption{Three fusion types of the connector in multi-modal scenario: (a) early fusion. (b) intermediate fusion. (c) late fusion.}
\label{f3}
\end{figure*}

\subsection{Multi-Layer Scenario}
Recent studies show that different layers of the same visual encoder capture distinct levels of abstraction and emphasize different regions of interest.
Taking \textit{CLIP} as an example, the shallow layer features focus on low-level details such as edges and textures, while the deep layer features are superior at high-level semantics such as objects and structures.
In the multi-layer scenario, the connector aims to effectively integrate features from multiple layers of a single encoder before performing its alignment role.
By leveraging the distinct strengths of different layers, ranging from low-level details to high-level semantics, the connector enhances feature representation through various fusion strategies, making alignment with textual embedding space more robust.
As shown in Table~\ref{t1}, we list the representative connectors in the multi-layer scenario according to the different properties.

LION~\citep{chen2024lion} proposes the vision aggregator which consists of two attention blocks to fuse the multi-level features from the selected [16, 32, 47] layers, facilitating the fine-grained knowledge learned based on visual grounding tasks.
In addition, LION also uses a Q-Former to process features of the penultimate layer and combines the results of two branches through token concatenation.
Dense Connector~\citep{yao2024dense} explores three intuitive instantiations for multi-layer fusion.
The first is sparse token integration (STI), which preserves the final layer features unchanged while downsampling additional visual features from other layers by using average pooling, followed by token concatenation.
The second is sparse channel integration (SCI), which directly concatenates features from the selected [8, 16, 24] layers along the channel dimension.
The third is dense channel integration (DCI), which partitions the features of all layers into groups, implements element-wise averaging within each group, and concatenates them with the last layer features along the channel dimension.
TokenPacker~\citep{li2024tokenpacker} utilizes the low-resolution interpolated visual features as point-based queries and injects the corresponding sub-region features of multiple layers into it using a point-to-region cross-attention operation.
In MMFuser~\citep{cao2024mmfuser}, deep [23] layer features are employed as query elements, while shallow and intermediate [3,8,13,18] layers features are concatenated to form key and value elements.

\subsection{Multi-Encoder Scenario}
Each encoder, trained on different datasets and optimized for specific tasks, provides a unique perspective on the input data.
For instance, \textit{CLIP}, pre-trained on text-image pairs using contrastive learning, excels in aligning textual and visual information, whereas \textit{DINO}, trained on pure images, performs exceptionally well in object detection tasks.
While the multi-layer scenario focuses on enhancing feature representations by combining different layers of a single encoder, the multi-encoder scenario takes this a step further by integrating more comprehensive and complementary features from multiple encoders. 
In the multi-encoder scenario, the connector aims to simultaneously achieve internal alignment of all encoder outputs and external alignment with the textual space.
As shown in Table~\ref{t2}, we list the representative connectors in the multi-encoder scenario according to the multi-encoder selection and fusion strategy.

As one of the pioneer works in combining complement multiple encoders, COMM~\citep{jiang2023clip} extracts features of all layers from \textit{CLIP} and deep layers from \textit{DINOv2}, then applies learnable coefficients to weight them separately before concatenating.
Eyes Wide Shut~\citep{tong2024eyes} utilizes a coefficient to control the portion of \textit{CLIP} and \textit{DINOv2} in the A-MoF connector, while interleaving two kinds of features along the token dimension in the I-MoF connector.
MoME~\citep{shen2024mome} uses an adaptive deformable transformation module to align the output of three encoders in the token and channel dimensions.
Then, based on the specific instructions of each sample, a soft router is applied to generate customized fusion ratios for different encoders.
Considering the demands of fine-grained visual information medical domain, LLaVA-Ultra~\citep{guo2024llava} extends visual encoders from \textit{CLIP} to \textit{SAM} and uses the weighted channel addition strategy to fuse the features.
Eagle~\citep{shi2024eagle} explores different combination schemes and fusion strategies in five visual encoders, and ultimately adopts direct channel concatenation due to its efficiency, scalability, and performance.
Cambrian-1~\citep{tong2024cambrian} partitions the feature maps of five visual encoders and facilitates the information aggregation via a set of learnable queries, taking the partitioned visual features of multiple encoders as keys and values.
BRAVE~\citep{kar2025brave} proposes MEQ-Former, which derives tokens from a set of learnable queries and takes the concatenated features from five different encoders as keys and values.

\subsection{Multi-Modal Scenario}
Driven by the abundance of image-text and video-text paired data, current MLLMs primarily focus on vision-text multi-modal tasks.
However, the true advance towards general-purpose AI demands the capability to handle a variety of modalities simultaneously.
Different modality-text pairs often exhibit significant differences in data distribution and representation spaces.
A key challenge lies in effectively fusing information from these diverse modalities into a cohesive, unified representation space.
In this multi-modal scenario, the connector is crucial:
if not designed properly, the integration of information from different modalities can lead to interference, ultimately undermining the model's performance. 

As illustrated in Figure~\ref{f3}, we summarize three fusion types of the connector in multi-modal scenario.
The first type is early fusion, which requires a powerful and robust multi-modal encoder capable of effectively aligning diverse modalities into a unified space before they are fed to the connector.
Therefore, the connector's task is relatively simple, as it only needs to align the unified space with the textual space.
For example, PandaGPT~\citep{su2023pandagpt} adopts ImageBind as the unified encoder and only uses the linear projection to align the representations.
The second type is intermediate fusion, which processes each modality through its own encoder before integrating the features at the connector. 
The connector should handle the challenging task of fusing and mapping features from different modalities, which often have significant disparities in their representation spaces, ensuring that each modality aligns well with the textual space.
MACAW-LLM~\citep{lyu2023macaw} and MEERKAT~\citep{chowdhury2024meerkat} are typical cases of intermediate fusion connectors through well-designed cross-modal attention mechanisms.
The third type is late fusion, which involves separate encoders and connectors for each modality.
Each connector aligns its respective modality with the textual space independently, and the aligned features are concatenated before being fed into the LLM.
AnyMAL~\citep{moon2024anymal} chooses resampler for image branch and linear layers for other modalities' branch.
GroundingGPT~\citep{li2024groundinggpt} applies MLP for image branch and Q-former for video and audio branch.
Note that CAT~\citep{ye2025cat} utilizes a combination of the three types mentioned above.

\section{Future Directions and Challenges}
As a crucial and effective component in the realization of MLLMs, studies on connector frameworks and applications are advancing rapidly, giving rise to numerous challenges and opportunities. 
We have identified several critical challenges and potential areas for future studies.

\paragraph{High-Resolution Input.}
To unlock the high-resolution image processing capabilities of MLLM and enhance fine-grained perception, current works such as InternVL 1.5~\citep{chen2024far} and HiRED~\citep{arif2024hired} typically partition high-resolution images into low-resolution patches and rely on general image encoders and common connectors for processing. 
However, pre-partitioning may disrupt spatial relationships, potentially distorting visual information. 
Therefore, rethinking how connectors handle relationships and interactions between different partitioned patches presents a potential optimization direction.

\paragraph{Dynamic Compression.}
Previous works usually compress tokens using a fixed number or rate, which lacks the flexibility to adapt to varying input complexities and contextual demands. 
This rigidity can lead to suboptimal performance, as fixed-length or fixed-rate compression may either discard critical information from rich inputs or introduce unnecessary redundancy for simple data.
FocusLLaVA~\citep{zhu2024focusllava} employs adaptive downsampling on different image patches based on local and global information, while DocKylin~\citep{zhang2024dockylin} utilizes K-means clustering to categorize and aggregate tokens through similarity-weighted summation.
There remains significant room for developing more adaptive compression mechanisms that can dynamically adjust to the unique characteristics of each input.

\paragraph{Guide Information Selection.}
Inspired by human cognition, where the brain selectively focuses on regions of interest based on guidance, connectors greatly benefit from guide information selection. 
This is exemplified in X-guided MoE connectors, which explore various guide information to determine expert weights.
Future connector designs should consider to facilitate deeper interactions between guide information and token representations.
For example, PPLLaVA~\citep{liu2024ppllava} employs a prompt-guided pooling module to extract video features relevant to user prompts.
World Knowledge~\citep{zhai2024world} proposes an instruction-guided interaction module to better align the visual features with the natural language instructions.
By selecting optimal guide information, connectors can focus on task-relevant features, alleviating the burden on the LLM to manage these interactions.

\paragraph{Combination Strategy.}
As discussed in Section 4, recent works have increasingly focused on using multi-layer or multi-encoder outputs to capture more comprehensive and multi-dimensional representations of the multi-modal information.
While these approaches enhance the richness of the feature representations, they also introduce the challenge of information redundancy.
Cambrian-1~\citep{tong2024cambrian} presents the SVA module to aggregate features from multiple visual encoders.
However, it requires repeated aggregation across multiple LLM layers to iteratively access and integrate necessary visual information. 
Eagle~\citep{shi2024eagle} has explored several combination strategies and surprisingly found that the simple channel concatenation was the most effective. 
These findings underscore the vast potential for improving combination strategies in connectors to balance the trade-off between representational richness and efficiency.

\paragraph{Interpretability.}
While the connector is pivotal in bridging the gap between modalities, its effectiveness has not been fully explored from an interpretability perspective.
Investigating interpretability can provide valuable insights into the inner mechanisms of the connector, guiding the design of more robust and efficient architectures while addressing potential weaknesses.
For instance, MMNeuron~\citep{huo2024mmneuron} analyzes the distribution of domain-specific neurons across different modules, suggesting that the number of such neurons reflects their understanding capabilities.
Deco~\citep{yao2024deco} uses R-GAE relevance maps to trace the flow of information from the generated textual tokens back to raw visual patches and intermediate connector output, revealing potential semantic deficiencies in the Q-former connector, such as the loss of fine-grained attribute and spatial locality.
We believe that future research on the interpretability of connectors will play a key role in unleashing the full potential of MLLMs.

\section{Conclusion}
The development of connectors in multi-modal large language models has become a critical area of research as the field progresses towards general AI.
In this survey, we aim to provide an in-depth overview of the connector in existing MLLMs.
Firstly, we introduce a new taxonomy that categorizes connectors into atomic operations and holistic designs.
Secondly, we systematically review the representative studies according to the taxonomy, highlighting the advancements in mapping, compression, and mixture of experts, as well as the holistic approaches for handling multi-layer, multi-encoder, and multi-modal scenarios.
Finally, we discuss some challenges and highlight several future research
directions. 
We hope this survey can serve as a useful reference for researchers, encouraging further exploration and innovation in the design of more efficient and robust connectors for MLLMs.

\section*{Acknowledgments}
This paper is supported by the Natural Science Foundation of Beijing with Grant No.25D30216.

\newpage
{
\fontsize{9}{10.5}\selectfont
\bibliographystyle{named}
\bibliography{cit}

\begin{thebibliography}{}

\bibitem[\protect\citeauthoryear{Alayrac \bgroup \em et al.\egroup }{2022}]{alayrac2022flamingo}
Jean-Baptiste Alayrac, Jeff Donahue, Pauline Luc, Antoine Miech, Iain Barr, Yana Hasson, Karel Lenc, Arthur Mensch, Katherine Millican, Malcolm Reynolds, et~al.
\newblock Flamingo: a visual language model for few-shot learning.
\newblock {\em Advances in neural information processing systems}, 35:23716--23736, 2022.

\bibitem[\protect\citeauthoryear{Arif \bgroup \em et al.\egroup }{2024}]{arif2024hired}
Kazi Hasan~Ibn Arif, JinYi Yoon, Dimitrios~S Nikolopoulos, Hans Vandierendonck, Deepu John, and Bo~Ji.
\newblock Hired: Attention-guided token dropping for efficient inference of high-resolution vision-language models in resource-constrained environments.
\newblock {\em arXiv preprint arXiv:2408.10945}, 2024.

\bibitem[\protect\citeauthoryear{Cao \bgroup \em et al.\egroup }{2024}]{cao2024mmfuser}
Yue Cao, Yangzhou Liu, Zhe Chen, Guangchen Shi, Wenhai Wang, Danhuai Zhao, and Tong Lu.
\newblock Mmfuser: Multimodal multi-layer feature fuser for fine-grained vision-language understanding.
\newblock {\em arXiv preprint arXiv:2410.11829}, 2024.

\bibitem[\protect\citeauthoryear{Cha \bgroup \em et al.\egroup }{2024}]{cha2024honeybee}
Junbum Cha, Wooyoung Kang, Jonghwan Mun, and Byungseok Roh.
\newblock Honeybee: Locality-enhanced projector for multimodal llm.
\newblock In {\em Proceedings of the IEEE/CVF Conference on Computer Vision and Pattern Recognition}, pages 13817--13827, 2024.

\bibitem[\protect\citeauthoryear{Chen \bgroup \em et al.\egroup }{2023}]{chen2023minigpt}
Jun Chen, Deyao Zhu, Xiaoqian Shen, Xiang Li, Zechun Liu, Pengchuan Zhang, Raghuraman Krishnamoorthi, Vikas Chandra, Yunyang Xiong, and Mohamed Elhoseiny.
\newblock Minigpt-v2: large language model as a unified interface for vision-language multi-task learning.
\newblock {\em arXiv preprint arXiv:2310.09478}, 2023.

\bibitem[\protect\citeauthoryear{Chen \bgroup \em et al.\egroup }{2024a}]{chen2024lion}
Gongwei Chen, Leyang Shen, Rui Shao, Xiang Deng, and Liqiang Nie.
\newblock Lion: Empowering multimodal large language model with dual-level visual knowledge.
\newblock In {\em Proceedings of the IEEE/CVF Conference on Computer Vision and Pattern Recognition}, pages 26540--26550, 2024.

\bibitem[\protect\citeauthoryear{Chen \bgroup \em et al.\egroup }{2024b}]{chen2024far}
Zhe Chen, Weiyun Wang, Hao Tian, Shenglong Ye, Zhangwei Gao, Erfei Cui, Wenwen Tong, Kongzhi Hu, Jiapeng Luo, Zheng Ma, et~al.
\newblock How far are we to gpt-4v? closing the gap to commercial multimodal models with open-source suites.
\newblock {\em Science China Information Sciences}, 67(12):220101, 2024.

\bibitem[\protect\citeauthoryear{Chen \bgroup \em et al.\egroup }{2024c}]{chen2024surgfc}
Zhen Chen, Xingjian Luo, Jinlin Wu, Danny~TM Chan, Zhen Lei, Sebastien Ourselin, and Hongbin Liu.
\newblock Surgfc: Multimodal surgical function calling framework on the demand of surgeons.
\newblock In {\em 2024 IEEE International Conference on Bioinformatics and Biomedicine (BIBM)}, pages 3076--3081. IEEE, 2024.

\bibitem[\protect\citeauthoryear{Chowdhury \bgroup \em et al.\egroup }{2024}]{chowdhury2024meerkat}
Sanjoy Chowdhury, Sayan Nag, Subhrajyoti Dasgupta, Jun Chen, Mohamed Elhoseiny, Ruohan Gao, and Dinesh Manocha.
\newblock Meerkat: Audio-visual large language model for grounding in space and time.
\newblock In {\em European Conference on Computer Vision}, pages 52--70. Springer, 2024.

\bibitem[\protect\citeauthoryear{Eom \bgroup \em et al.\egroup }{2024}]{eom2024query}
SooHwan Eom, Jay Shim, Gwanhyeong Koo, Haebin Na, Mark Hasegawa-Johnson, Sungwoong Kim, and Chang Yoo.
\newblock Query-based cross-modal projector bolstering mamba multimodal llm.
\newblock In {\em Findings of the Association for Computational Linguistics: EMNLP 2024}, pages 14158--14167, 2024.

\bibitem[\protect\citeauthoryear{Fei \bgroup \em et al.\egroup }{2024}]{fei2024vitron}
Hao Fei, Shengqiong Wu, Hanwang Zhang, Tat-Seng Chua, and Shuicheng Yan.
\newblock Vitron: A unified pixel-level vision llm for understanding, generating, segmenting, editing.
\newblock In {\em Proceedings of the Advances in neural information processing systems}, 2024.

\bibitem[\protect\citeauthoryear{Guo \bgroup \em et al.\egroup }{2024}]{guo2024llava}
Xuechen Guo, Wenhao Chai, Shi-Yan Li, and Gaoang Wang.
\newblock Llava-ultra: Large chinese language and vision assistant for ultrasound.
\newblock In {\em Proceedings of the 32nd ACM International Conference on Multimedia}, pages 8845--8854, 2024.

\bibitem[\protect\citeauthoryear{Han \bgroup \em et al.\egroup }{2024}]{han2024onellm}
Jiaming Han, Kaixiong Gong, Yiyuan Zhang, Jiaqi Wang, Kaipeng Zhang, Dahua Lin, Yu~Qiao, Peng Gao, and Xiangyu Yue.
\newblock Onellm: One framework to align all modalities with language.
\newblock In {\em Proceedings of the IEEE/CVF Conference on Computer Vision and Pattern Recognition}, pages 26584--26595, 2024.

\bibitem[\protect\citeauthoryear{Huo \bgroup \em et al.\egroup }{2024}]{huo2024mmneuron}
Jiahao Huo, Yibo Yan, Boren Hu, Yutao Yue, and Xuming Hu.
\newblock Mmneuron: Discovering neuron-level domain-specific interpretation in multimodal large language model.
\newblock {\em arXiv preprint arXiv:2406.11193}, 2024.

\bibitem[\protect\citeauthoryear{Jiang \bgroup \em et al.\egroup }{2023}]{jiang2023clip}
Dongsheng Jiang, Yuchen Liu, Songlin Liu, Jin'e Zhao, Hao Zhang, Zhen Gao, Xiaopeng Zhang, Jin Li, and Hongkai Xiong.
\newblock From clip to dino: Visual encoders shout in multi-modal large language models.
\newblock {\em arXiv preprint arXiv:2310.08825}, 2023.

\bibitem[\protect\citeauthoryear{Kar \bgroup \em et al.\egroup }{2025}]{kar2025brave}
O{\u{g}}uzhan~Fatih Kar, Alessio Tonioni, Petra Poklukar, Achin Kulshrestha, Amir Zamir, and Federico Tombari.
\newblock Brave: Broadening the visual encoding of vision-language models.
\newblock In {\em European Conference on Computer Vision}, pages 113--132. Springer, 2025.

\bibitem[\protect\citeauthoryear{Li \bgroup \em et al.\egroup }{2023}]{li2023blip}
Junnan Li, Dongxu Li, Silvio Savarese, and Steven Hoi.
\newblock Blip-2: Bootstrapping language-image pre-training with frozen image encoders and large language models.
\newblock In {\em International conference on machine learning}, pages 19730--19742. PMLR, 2023.

\bibitem[\protect\citeauthoryear{Li \bgroup \em et al.\egroup }{2024a}]{li2024tokenpacker}
Wentong Li, Yuqian Yuan, Jian Liu, Dongqi Tang, Song Wang, Jie Qin, Jianke Zhu, and Lei Zhang.
\newblock Tokenpacker: Efficient visual projector for multimodal llm.
\newblock {\em arXiv preprint arXiv:2407.02392}, 2024.

\bibitem[\protect\citeauthoryear{Li \bgroup \em et al.\egroup }{2024b}]{li20243dmit}
Zeju Li, Chao Zhang, Xiaoyan Wang, Ruilong Ren, Yifan Xu, Ruifei Ma, Xiangde Liu, and Rong Wei.
\newblock 3dmit: 3d multi-modal instruction tuning for scene understanding.
\newblock In {\em 2024 IEEE International Conference on Multimedia and Expo Workshops (ICMEW)}, pages 1--5. IEEE, 2024.

\bibitem[\protect\citeauthoryear{Li \bgroup \em et al.\egroup }{2024c}]{li2024groundinggpt}
Zhaowei Li, Qi~Xu, Dong Zhang, Hang Song, Yiqing Cai, Qi~Qi, Ran Zhou, Junting Pan, Zefeng Li, Vu~Tu, et~al.
\newblock Groundinggpt: Language enhanced multi-modal grounding model.
\newblock In {\em Proceedings of the 62nd Annual Meeting of the Association for Computational Linguistics (Volume 1: Long Papers)}, pages 6657--6678, 2024.

\bibitem[\protect\citeauthoryear{Li \bgroup \em et al.\egroup }{2025}]{li2025cumo}
Jiachen Li, Xinyao Wang, Sijie Zhu, Chia-Wen Kuo, Lu~Xu, Fan Chen, Jitesh Jain, Humphrey Shi, and Longyin Wen.
\newblock Cumo: Scaling multimodal llm with co-upcycled mixture-of-experts.
\newblock {\em Advances in Neural Information Processing Systems}, 37:131224--131246, 2025.

\bibitem[\protect\citeauthoryear{Liu \bgroup \em et al.\egroup }{2024a}]{liu2024infimm}
Haogeng Liu, Quanzeng You, Yiqi Wang, Xiaotian Han, Bohan Zhai, Yongfei Liu, Wentao Chen, Yiren Jian, Yunzhe Tao, Jianbo Yuan, et~al.
\newblock Infimm: Advancing multimodal understanding with an open-sourced visual language model.
\newblock In {\em Findings of the Association for Computational Linguistics ACL 2024}, pages 485--492, 2024.

\bibitem[\protect\citeauthoryear{Liu \bgroup \em et al.\egroup }{2024b}]{liu2024improved}
Haotian Liu, Chunyuan Li, Yuheng Li, and Yong~Jae Lee.
\newblock Improved baselines with visual instruction tuning.
\newblock In {\em Proceedings of the IEEE/CVF Conference on Computer Vision and Pattern Recognition}, pages 26296--26306, 2024.

\bibitem[\protect\citeauthoryear{Liu \bgroup \em et al.\egroup }{2024c}]{liu2024visual}
Haotian Liu, Chunyuan Li, Qingyang Wu, and Yong~Jae Lee.
\newblock Visual instruction tuning.
\newblock {\em Advances in neural information processing systems}, 36, 2024.

\bibitem[\protect\citeauthoryear{Liu \bgroup \em et al.\egroup }{2024d}]{liu2024ppllava}
Ruyang Liu, Haoran Tang, Haibo Liu, Yixiao Ge, Ying Shan, Chen Li, and Jiankun Yang.
\newblock Ppllava: Varied video sequence understanding with prompt guidance.
\newblock {\em arXiv preprint arXiv:2411.02327}, 2024.

\bibitem[\protect\citeauthoryear{Lyu \bgroup \em et al.\egroup }{2023}]{lyu2023macaw}
Chenyang Lyu, Minghao Wu, Longyue Wang, Xinting Huang, Bingshuai Liu, Zefeng Du, Shuming Shi, and Zhaopeng Tu.
\newblock Macaw-llm: Multi-modal language modeling with image, audio, video, and text integration.
\newblock {\em arXiv preprint arXiv:2306.09093}, 2023.

\bibitem[\protect\citeauthoryear{McKinzie \bgroup \em et al.\egroup }{2025}]{mckinzie2025mm1}
Brandon McKinzie, Zhe Gan, Jean-Philippe Fauconnier, Sam Dodge, Bowen Zhang, Philipp Dufter, Dhruti Shah, Xianzhi Du, Futang Peng, Anton Belyi, et~al.
\newblock Mm1: methods, analysis and insights from multimodal llm pre-training.
\newblock In {\em European Conference on Computer Vision}, pages 304--323. Springer, 2025.

\bibitem[\protect\citeauthoryear{Moon \bgroup \em et al.\egroup }{2024}]{moon2024anymal}
Seungwhan Moon, Andrea Madotto, Zhaojiang Lin, Tushar Nagarajan, Matt Smith, Shashank Jain, Chun-Fu Yeh, Prakash Murugesan, Peyman Heidari, Yue Liu, et~al.
\newblock Anymal: An efficient and scalable any-modality augmented language model.
\newblock In {\em Proceedings of the 2024 Conference on Empirical Methods in Natural Language Processing: Industry Track}, pages 1314--1332, 2024.

\bibitem[\protect\citeauthoryear{Shen \bgroup \em et al.\egroup }{2024}]{shen2024mome}
Leyang Shen, Gongwei Chen, Rui Shao, Weili Guan, and Liqiang Nie.
\newblock Mome: Mixture of multimodal experts for generalist multimodal large language models.
\newblock {\em arXiv preprint arXiv:2407.12709}, 2024.

\bibitem[\protect\citeauthoryear{Shi \bgroup \em et al.\egroup }{2024}]{shi2024eagle}
Min Shi, Fuxiao Liu, Shihao Wang, Shijia Liao, Subhashree Radhakrishnan, De-An Huang, Hongxu Yin, Karan Sapra, Yaser Yacoob, Humphrey Shi, et~al.
\newblock Eagle: Exploring the design space for multimodal llms with mixture of encoders.
\newblock {\em arXiv preprint arXiv:2408.15998}, 2024.

\bibitem[\protect\citeauthoryear{Song \bgroup \em et al.\egroup }{2023}]{song2023bridge}
Shezheng Song, Xiaopeng Li, Shasha Li, Shan Zhao, Jie Yu, Jun Ma, Xiaoguang Mao, and Weimin Zhang.
\newblock How to bridge the gap between modalities: A comprehensive survey on multimodal large language model.
\newblock {\em arXiv preprint arXiv:2311.07594}, 2023.

\bibitem[\protect\citeauthoryear{Su \bgroup \em et al.\egroup }{2023}]{su2023pandagpt}
Yixuan Su, Tian Lan, Huayang Li, Jialu Xu, Yan Wang, and Deng Cai.
\newblock Pandagpt: One model to instruction-follow them all.
\newblock {\em arXiv preprint arXiv:2305.16355}, 2023.

\bibitem[\protect\citeauthoryear{Tong \bgroup \em et al.\egroup }{2024a}]{tong2024cambrian}
Shengbang Tong, Ellis Brown, Penghao Wu, Sanghyun Woo, Manoj Middepogu, Sai~Charitha Akula, Jihan Yang, Shusheng Yang, Adithya Iyer, Xichen Pan, et~al.
\newblock Cambrian-1: A fully open, vision-centric exploration of multimodal llms.
\newblock {\em arXiv preprint arXiv:2406.16860}, 2024.

\bibitem[\protect\citeauthoryear{Tong \bgroup \em et al.\egroup }{2024b}]{tong2024eyes}
Shengbang Tong, Zhuang Liu, Yuexiang Zhai, Yi~Ma, Yann LeCun, and Saining Xie.
\newblock Eyes wide shut? exploring the visual shortcomings of multimodal llms.
\newblock In {\em Proceedings of the IEEE/CVF Conference on Computer Vision and Pattern Recognition}, pages 9568--9578, 2024.

\bibitem[\protect\citeauthoryear{Wang \bgroup \em et al.\egroup }{2024a}]{wang2024pargo}
An-Lan Wang, Bin Shan, Wei Shi, Kun-Yu Lin, Xiang Fei, Guozhi Tang, Lei Liao, Jingqun Tang, Can Huang, and Wei-Shi Zheng.
\newblock Pargo: Bridging vision-language with partial and global views.
\newblock {\em arXiv preprint arXiv:2408.12928}, 2024.

\bibitem[\protect\citeauthoryear{Wang \bgroup \em et al.\egroup }{2024b}]{wang2024q}
Hanzi Wang, Jiamin Ren, Yifeng Ding, Lei Ren, Huixing Jiang, Wei Chen, Fangxiang Feng, and Xiaojie Wang.
\newblock Q-moe: Connector for mllms with text-driven routing.
\newblock In {\em Proceedings of the 32nd ACM International Conference on Multimedia}, pages 817--825, 2024.

\bibitem[\protect\citeauthoryear{Wu and Xie}{2024}]{wu2024v}
Penghao Wu and Saining Xie.
\newblock V$^*$: Guided visual search as a core mechanism in multimodal llms.
\newblock In {\em Proceedings of the IEEE/CVF Conference on Computer Vision and Pattern Recognition}, pages 13084--13094, 2024.

\bibitem[\protect\citeauthoryear{Xu \bgroup \em et al.\egroup }{2024a}]{xu2024pllava}
Lin Xu, Yilin Zhao, Daquan Zhou, Zhijie Lin, See~Kiong Ng, and Jiashi Feng.
\newblock Pllava: Parameter-free llava extension from images to videos for video dense captioning.
\newblock {\em arXiv preprint arXiv:2404.16994}, 2024.

\bibitem[\protect\citeauthoryear{Xu \bgroup \em et al.\egroup }{2024b}]{xu2024chartmoe}
Zhengzhuo Xu, Bowen Qu, Yiyan Qi, Sinan Du, Chengjin Xu, Chun Yuan, and Jian Guo.
\newblock Chartmoe: Mixture of expert connector for advanced chart understanding.
\newblock {\em arXiv preprint arXiv:2409.03277}, 2024.

\bibitem[\protect\citeauthoryear{Yan \bgroup \em et al.\egroup }{2023}]{yan2023voila}
Kun Yan, Lei Ji, Zeyu Wang, Yuntao Wang, Nan Duan, and Shuai Ma.
\newblock Voila-a: Aligning vision-language models with user's gaze attention.
\newblock {\em arXiv preprint arXiv:2401.09454}, 2023.

\bibitem[\protect\citeauthoryear{Yang \bgroup \em et al.\egroup }{2024}]{yang2024planllm}
Dejie Yang, Zijing Zhao, et~al.
\newblock Planllm: Video procedure planning with refinable large language models.
\newblock {\em arXiv preprint arXiv:2412.19139}, 2024.

\bibitem[\protect\citeauthoryear{Yao \bgroup \em et al.\egroup }{2024a}]{yao2024dense}
Huanjin Yao, Wenhao Wu, Taojiannan Yang, YuXin Song, Mengxi Zhang, Haocheng Feng, Yifan Sun, Zhiheng Li, Wanli Ouyang, and Jingdong Wang.
\newblock Dense connector for mllms.
\newblock {\em arXiv preprint arXiv:2405.13800}, 2024.

\bibitem[\protect\citeauthoryear{Yao \bgroup \em et al.\egroup }{2024b}]{yao2024deco}
Linli Yao, Lei Li, Shuhuai Ren, Lean Wang, Yuanxin Liu, Xu~Sun, and Lu~Hou.
\newblock Deco: Decoupling token compression from semantic abstraction in multimodal large language models.
\newblock {\em arXiv preprint arXiv:2405.20985}, 2024.

\bibitem[\protect\citeauthoryear{Ye \bgroup \em et al.\egroup }{2024}]{ye2024mplug}
Jiabo Ye, Haiyang Xu, Haowei Liu, Anwen Hu, Ming Yan, Qi~Qian, Ji~Zhang, Fei Huang, and Jingren Zhou.
\newblock mplug-owl3: Towards long image-sequence understanding in multi-modal large language models.
\newblock {\em arXiv preprint arXiv:2408.04840}, 2024.

\bibitem[\protect\citeauthoryear{Ye \bgroup \em et al.\egroup }{2025}]{ye2025cat}
Qilang Ye, Zitong Yu, Rui Shao, Xinyu Xie, Philip Torr, and Xiaochun Cao.
\newblock Cat: Enhancing multimodal large language model to answer questions in dynamic audio-visual scenarios.
\newblock In {\em European Conference on Computer Vision}, pages 146--164. Springer, 2025.

\bibitem[\protect\citeauthoryear{Yin \bgroup \em et al.\egroup }{2023}]{yin2023survey}
Shukang Yin, Chaoyou Fu, Sirui Zhao, Ke~Li, Xing Sun, Tong Xu, and Enhong Chen.
\newblock A survey on multimodal large language models.
\newblock {\em arXiv preprint arXiv:2306.13549}, 2023.

\bibitem[\protect\citeauthoryear{Young \bgroup \em et al.\egroup }{2024}]{young2024yi}
Alex Young, Bei Chen, Chao Li, Chengen Huang, Ge~Zhang, Guanwei Zhang, Heng Li, Jiangcheng Zhu, Jianqun Chen, Jing Chang, et~al.
\newblock Yi: Open foundation models by 01. ai.
\newblock {\em arXiv preprint arXiv:2403.04652}, 2024.

\bibitem[\protect\citeauthoryear{Zhai \bgroup \em et al.\egroup }{2024}]{zhai2024world}
Mingliang Zhai, Cheng Li, Zengyuan Guo, Ningrui Yang, Xiameng Qin, Yuwei Wu, Sanyuan Zhao, Junyu Han, Ji~Tao, and Yunde Jia.
\newblock World knowledge-enhanced reasoning using instruction-guided interactor in autonomous driving.
\newblock {\em arXiv preprint arXiv:2412.06324}, 2024.

\bibitem[\protect\citeauthoryear{Zhang \bgroup \em et al.\egroup }{2024a}]{zhang2024mm}
Duzhen Zhang, Yahan Yu, Jiahua Dong, Chenxing Li, Dan Su, Chenhui Chu, and Dong Yu.
\newblock Mm-llms: Recent advances in multimodal large language models.
\newblock {\em arXiv preprint arXiv:2401.13601}, 2024.

\bibitem[\protect\citeauthoryear{Zhang \bgroup \em et al.\egroup }{2024b}]{zhang2024dockylin}
Jiaxin Zhang, Wentao Yang, Songxuan Lai, Zecheng Xie, and Lianwen Jin.
\newblock Dockylin: A large multimodal model for visual document understanding with efficient visual slimming.
\newblock {\em arXiv preprint arXiv:2406.19101}, 2024.

\bibitem[\protect\citeauthoryear{Zhu \bgroup \em et al.\egroup }{2023}]{zhu2023minigpt}
Deyao Zhu, Jun Chen, Xiaoqian Shen, Xiang Li, and Mohamed Elhoseiny.
\newblock Minigpt-4: Enhancing vision-language understanding with advanced large language models.
\newblock {\em arXiv preprint arXiv:2304.10592}, 2023.

\bibitem[\protect\citeauthoryear{Zhu \bgroup \em et al.\egroup }{2024a}]{zhu2024uni}
Xun Zhu, Ying Hu, Fanbin Mo, Miao Li, and Ji~Wu.
\newblock Uni-med: A unified medical generalist foundation model for multi-task learning via connector-moe.
\newblock {\em arXiv preprint arXiv:2409.17508}, 2024.

\bibitem[\protect\citeauthoryear{Zhu \bgroup \em et al.\egroup }{2024b}]{zhu2024focusllava}
Yuke Zhu, Chi Xie, Shuang Liang, Bo~Zheng, and Sheng Guo.
\newblock Focusllava: A coarse-to-fine approach for efficient and effective visual token compression.
\newblock {\em arXiv preprint arXiv:2411.14228}, 2024.

\end{thebibliography}
}

\end{document}